\documentclass{Interspeech}

\interspeechcameraready

\title{Automatic classification of stop realisation with wav2vec2.0}

\author[affiliation={1}]{James}{Tanner}
\author[affiliation={2}]{Morgan}{Sonderegger}
\author[affiliation={1}]{Jane}{Stuart-Smith}
\author[affiliation={3}]{Jeff}{Mielke}
\author[affiliation={4}]{Tyler}{Kendall}

\affiliation{Department of English Language and Linguistics}{University of Glasgow}{United Kingdom}
\affiliation{Department of Linguistics}{McGill University}{Canada}
\affiliation{Department of English}{North Carolina State University}{United States}
\affiliation{School of Law}{Duke University}{United States}
\email{\{james.tanner,jane.stuart-smith\}@glasgow.ac.uk, morgan.sonderegger@mcgill.ca, jimielke@ncsu.edu, tyler.kendall@law.duke.edu}


\begin{document}

\maketitle

\begin{abstract}
Modern phonetic research regularly makes use of automatic tools for the annotation of speech data, however few tools exist for the annotation of many variable phonetic phenomena. At the same time, pre-trained self-supervised models, such as wav2vec2.0, have been shown to perform well at speech classification tasks and latently encode fine-grained phonetic information. We demonstrate that wav2vec2.0 models can be trained to automatically classify stop burst presence with high accuracy in both English and Japanese, robust across both finely-curated and unprepared speech corpora. Patterns of variability in stop realisation are replicated with the automatic annotations, and closely follow those of manual annotations. These results demonstrate the potential of pre-trained speech models as tools for the automatic annotation and processing of speech corpus data, enabling researchers to `scale-up' the scope of phonetic research with relative ease.
\end{abstract}

\section{Introduction}

Over the past twenty years, a number of tools for the automatic and semi-automatic annotation of speech data have been released, including for the time-alignment of phonetic transcriptions to audio (`forced aligners') \cite{mfa,fave,labbcat} and the labelling of variable phonetic phenomena \cite{villareal2020,kendall2021,kim24l_interspeech}, both substantially reducing the time required for annotating speech data and greatly increasing the possible scale of phonetic research \cite{liberman2018}. Given the increasing size and complexity of speech datasets, it is equally important that these tools are both highly accurate and robust to variability, including cross-linguistic and cross-dialectal differences, as well as differences in recording quality and the degree of manual correction in the corpus development process.

\begin{figure*}[t]
  \centering
  \includegraphics[width=0.33\linewidth]{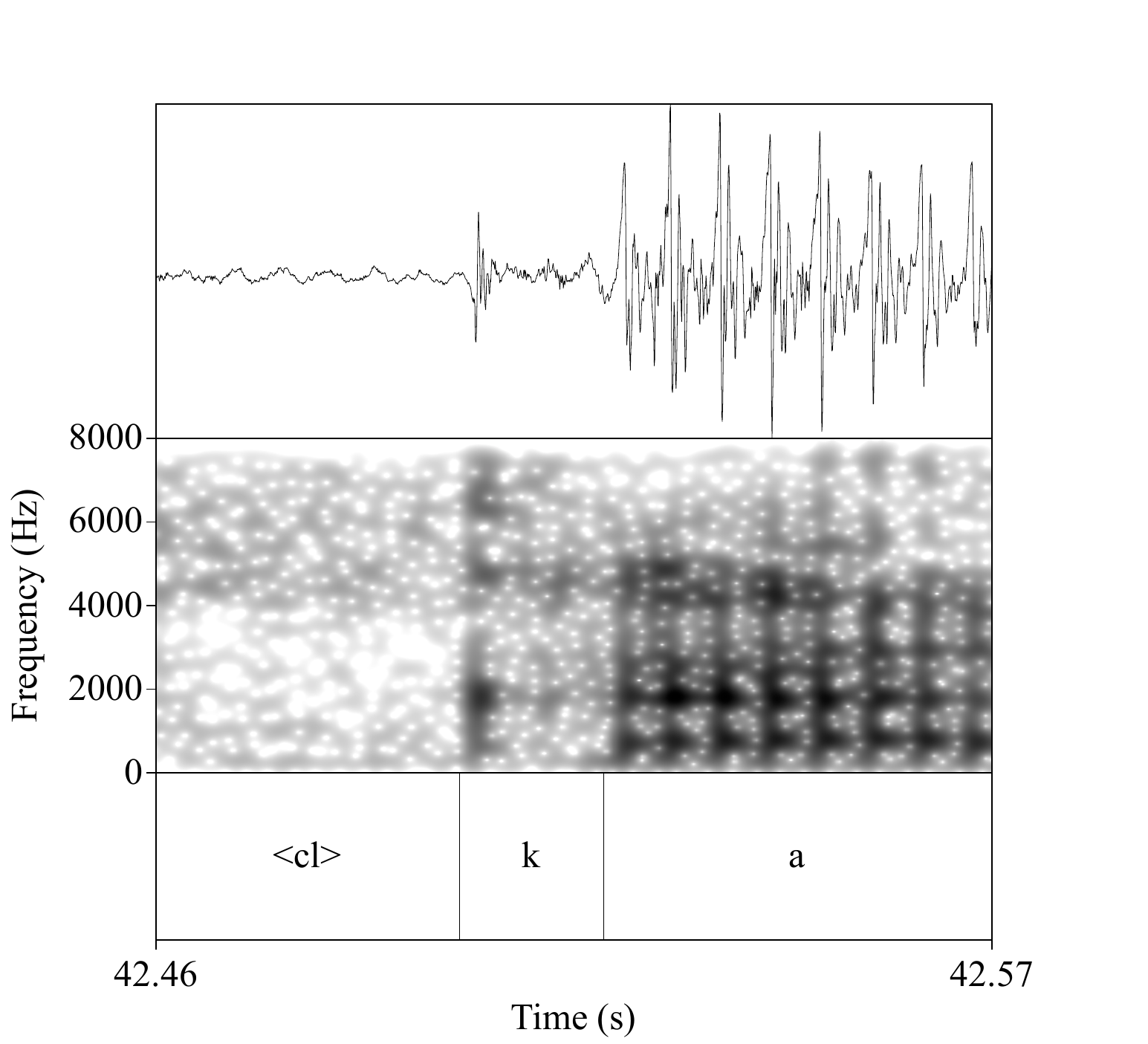}
  \includegraphics[width=0.33\linewidth]{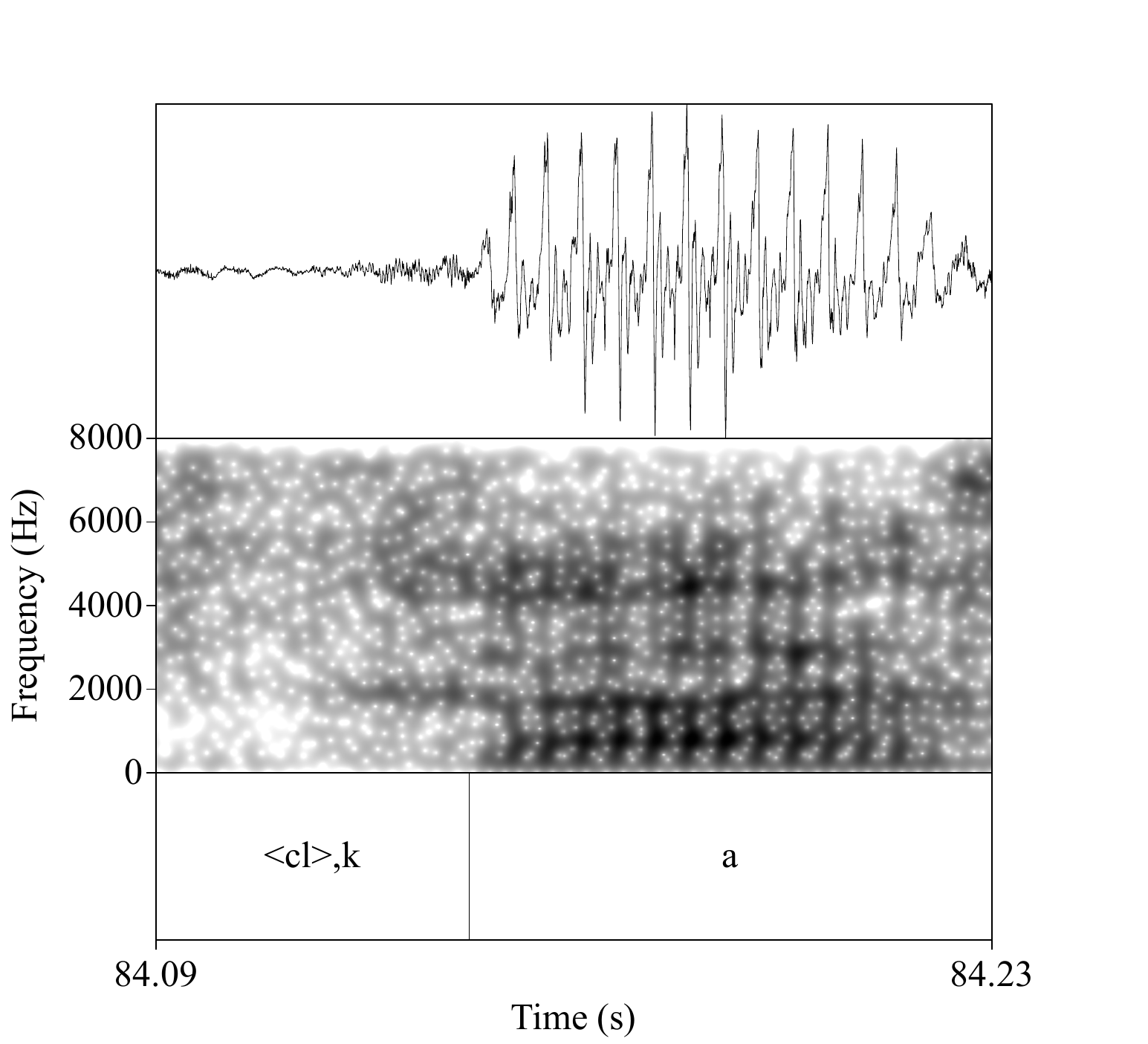}
  \includegraphics[width=0.33\linewidth]{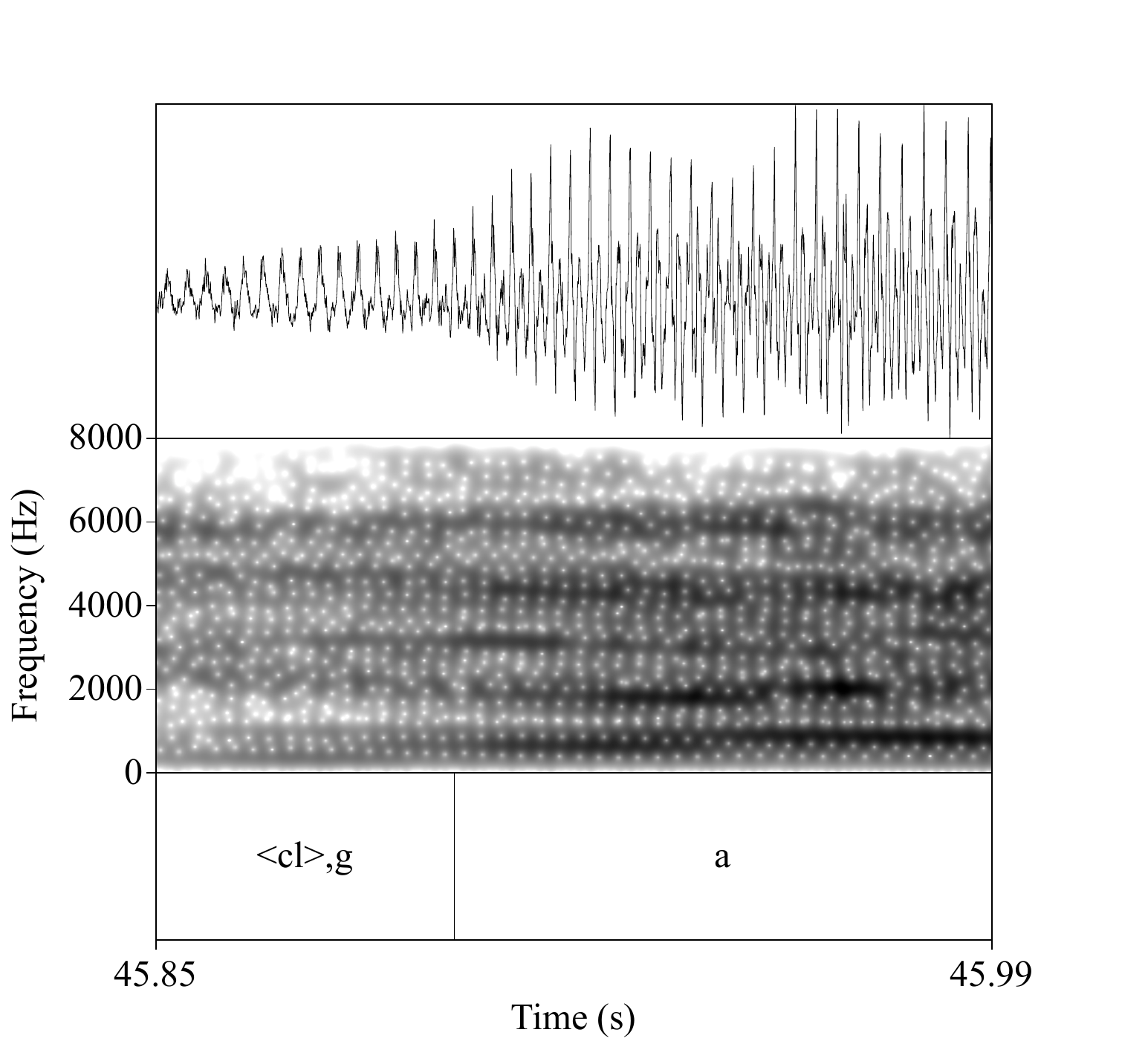}
  \caption{Waveforms, spectrograms, and TextGrid annotations for three utterance-medial stops from the CSJ-C (Speaker 459), showing a fully-realised stop with closure and release (left), a stop with a fricativised realisation (middle), and an incomplete closure with voicing continuing through the closure portion (right). TextGrid annotations reflect the binary distinction between a `realised' and `unrealised' burst, based on the \texttt{<cl>} and \texttt{phone} labels annotated as separate or combined (e.g. \texttt{<cl>,g}) respectively.}
  \label{fig:csj-specs}
\end{figure*}

One such kind of variation is that of the variable realisation of stop consonants. Stops are complex phonetic events containing multiple `stages', including the cessation of spectral activity from the closure of the oral cavity, a subsequent spike in energy from the closure release (the `burst'), and the offset of the burst into the following segment (Fig.~\ref{fig:csj-specs} left). Stops can vary in their qualitative (`allophonic') realisation, fundamentally characterised by the presence or absence of the stop burst (Fig.~\ref{fig:csj-specs} left vs. middle, right])), as well as in quantitative differences (e.g. voice onset time). These patterns of stop realisation have been of substantial interest from many different perspectives, including across languages \cite{Bouavichith2013,dicanio2022} and speech styles \cite{lavoie2001,warnertucker2011}.

Given the widespread focus on the phonetics of stop realisation, there is in turn interest in tools for the automatic annotation of stops \cite{autovot-program,shrem2019dr}. For the classification of allophonic stop realisation, \cite{dicanio2022} demonstrated that Mixtec voiceless stop realisation (stop vs fricative) can be predicted with high accuracy (98\%) in a manually-corrected speech corpus. We are not aware, however, of any such tool for stop realisation classification that is broadly available to the speech community, including one that performs well across languages, speech styles, and differences in corpus preparation. At the same time, recent large pre-trained neural network models -- such as Whisper \cite{whisper2022}, WavLM \cite{wavlm}, and wav2vec \cite{wav2vec,wav2vec2} -- have been shown to out-perform previous benchmarks in a range of speech-related tasks including speech recognition and speaker diarisation \cite{kunesova2024}. wav2vec2.0, a self-supervised convolutional and transformer network architecture, encodes a high-dimensional feature space from input waveforms, which can then be trained for subsequent downstream classification and regression tasks. Recent research has demonstrated that wav2vec2.0 featural space latently encodes fine-grained phonetic information \cite{dieck2022,choi2022,deheerkloots2024} and has also been used for the annotation of variable nasality \cite{kim24l_interspeech}. The question follows, then, whether a large pre-trained speech model such as wav2vec2.0 can be trained to automatically detect the variable realisation of stops in speech corpora which vary in recording quality and degree of manual curation.

Specifically, the research questions of this study are: \emph{How well can wav2vec2.0 classify the presence/absence of a stop burst in both 1. clean manually-corrected data and 2. noisy, largely uncorrected data?} (RQ1), \emph{How much data is needed to achieve good predictive performance?} (RQ2), and \emph{How do the patterns of stop realisation compare between predicted and hand-annotated stops?} (RQ3). RQ1 and RQ2 are addressed through two model training experiments, where different base models are trained to classify stop realisation with a corpus of spontaneous Japanese speech (Sec.~\ref{sec:exp1}) and a large multi-corpus dataset of English speech (Sec.~\ref{sec:exp2}). RQ3 is addressed through a comparison from the trained models to those from manual annotations, particularly with respect to patterns across phonological voicing and stop duration (Sec.~\ref{sec:analysis}).

\section{Experiment 1}
\label{sec:exp1}

The goal of Experiment 1 is to test whether wav2vec2.0 can be trained to accurately predict the presence or absence of a stop closure and burst in the `best-case' spontaneous speech context: speech that is cleanly-recorded and stylistically largely homogenous, and whose segmental annotations have been hand-corrected after alignment. 

\subsection{Data}
The data for Experiment 1 comes from the Core section of the \emph{Corpus of Spontaneous Japanese} (CSJ-C) \cite{Maekawa2000}, containing around 45 hours of Japanese speech (predominantly monologues) from 137 speakers. The CSJ-C contains extensive manual annotation \cite{csj-labelling,maekawa2018}, including the hand-correction of segmental boundaries and the presence or absence of bursts within a stop. Stops with distinct closure and burst portions are marked as separate \texttt{<cl>} and \texttt{phone} labels (e.g. Fig.~\ref{fig:csj-specs} left); where separate closure and bursts are not observable from the waveform or spectrogram, these are marked as a single \texttt{<cl>,phone} label (e.g. Fig.~\ref{fig:csj-specs}, middle \& left). Metadata (token ID, stop start time, stop end time, burst presence/absence) for 189,767 stops (without previous/following devoiced vowels) were extracted from the CSJ-C \cite{koiso2014}. To create the dataset, the audio for these stops were extracted from the corpus with an additional 10ms context window prepended/appended to the start and end times respectively. To avoid model overfitting, a random subset of 40k stops (20k voiced and voiceless stops) were selected as the final training set.

\subsection{Results}

\begin{table}[t]
  \caption{Number of model parameters, pre-training data, and accuracy and F1 metrics for each wav2vec2.0 base model (finetuned with 80\% of the CSJ-C data) on the 20\% test CSJ-C set. LS = Librispeech, LL = Librilight, MLS = Multilingual Librispeech, CV = CommonVoice.}
  \label{tab:csj-metrics}
  \centering
  \setlength\tabcolsep{.5\tabcolsep}
  \begin{tabular}{lllrr}
  \toprule
  \textbf{Model} & \textbf{N params} & \textbf{Pre-training data} & \textbf{Acc} & \textbf{F1} \\
  \midrule
  \texttt{base-960h} & 95M & LS & $0.94$ & $0.9$ \\ 
  \texttt{large-960h} & 317M & LS & $0.91$ & $0.86$ \\ 
  \texttt{lv60} & 317M & LS, LL & $0.94$ & $0.9$ \\ 
  \texttt{xslr} & 317M & MLS, CV, BABEL & $0.94$ & $0.9$ \\ 
  \bottomrule
  \end{tabular}
\end{table}

This dataset (split into 80\%-20\% train-test set) was used to train (as in \cite{kunesova2024}) stop burst presence/absence classifiers using four different wav2vec2.0 models -- three pre-trained on English data (\{\texttt{base-960h}, \texttt{large-960h}, \texttt{lv60}\}) and one trained on multilingual data (\{\texttt{xlsr}\}) -- using the HuggingFace \texttt{Transformers} library v4.44 \cite{wolf2020} in Python 3.11. Each model was trained on a 40GB NVIDIA A100 GPU for 10 epochs with a batch size of 64 for both training and evaluation, with a cosine learning rate schedule and a warmup ratio of 0.1. Training took between 10 minutes (\texttt{base-960h}) and 30 minutes (\texttt{lv60}). The test accuracy and F1 scores for each of the four models trained on CSJ-C data are shown in Table \ref{tab:csj-metrics}. The predictive accuracies of these models aligns closely to the $\sim$97\% reported for a previous burst detection model on similarly-corrected data \cite{dicanio2022}, indicating that wav2vec2.0 can be successfully trained for predicting the variable phonetic realisation of stops, at least in manually-curated speech corpora.

\section{Experiment 2}
\label{sec:exp2}

\begin{figure*}[t]
  \centering
  \includegraphics[width=\linewidth]{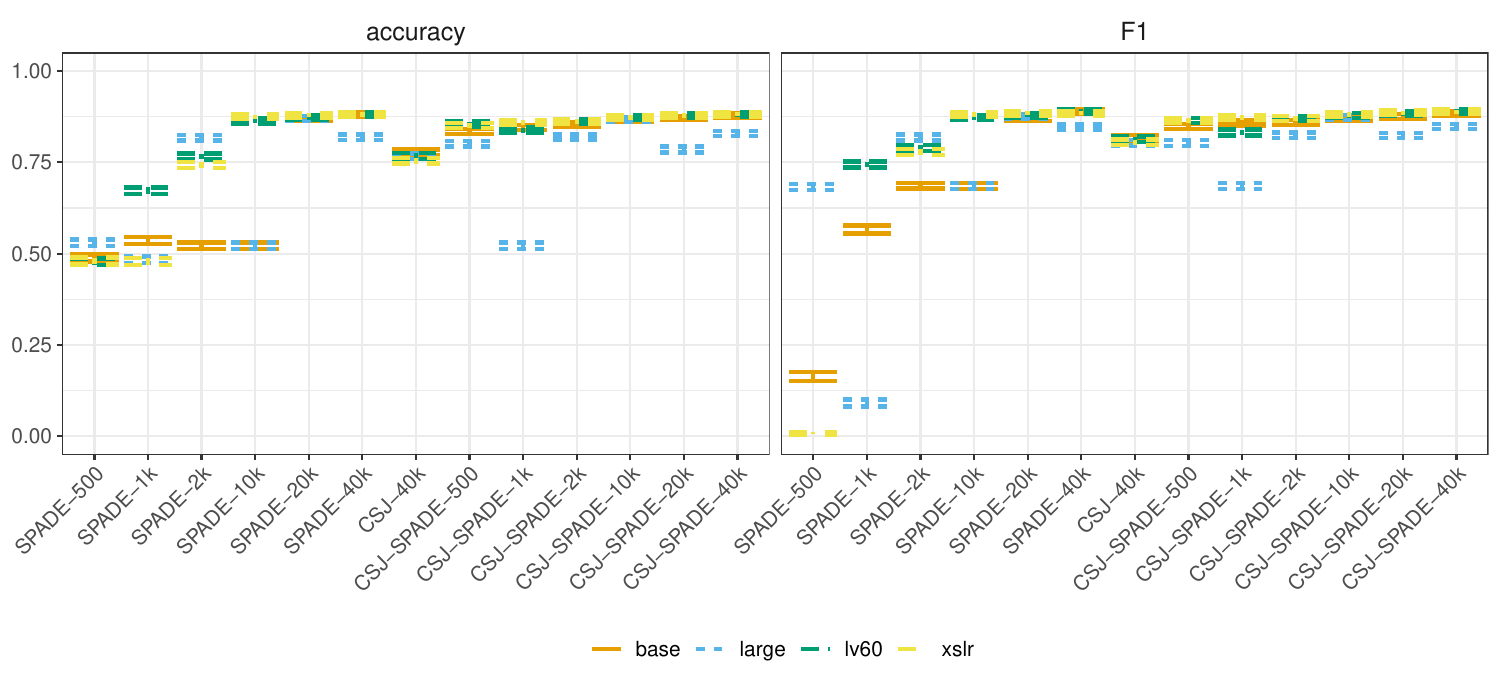}
  \caption{Accuracy and F1 metrics for wav2vec2.0 models on the 11k validation dataset with 95\% BCa bootstrapped confidence intervals \cite{boot}. X-axis corresponds to the CSJ-C models (without further training), the models trained using only the SPADE data, and the CSJ-C models that were subsequently trained with SPADE data.}
  \label{fig:ci_plot}
\end{figure*}

Having demonstrated that high predictive performance can be achieved with clean, manually-corrected speech data, Experiment 2 explores the performance of these models on more variable speech -- both in speech style and recording conditions -- with no manual correction following forced alignment.

\subsection{Data}

The data for Experiment 2 comes from \emph{The SPeech Across Dialects of English} (SPADE) project \cite{dmuc}, a large multi-corpus dataset of English speech, comprising of more than 40 corpora and 1,500 speakers (approximately 2,000 hours). The corpora comprising this dataset are themselves highly variable in both time and space, reflecting a range of British and North American dialects recorded between the 1970s and 2010s. Stylistically, this dataset contains speech collected from a range of recording contexts, including sociolinguistic interviews, spontaneous conversations, and reading passages. All of the corpora were force-aligned and these generated segment labels and boundaries were not subsequently corrected.

As this dataset does not contain labels for burst presence/absence, a training subset was manually annotated. As with Experiment 1, 1.13 million stops in the dataset was extracted: to aid in manual annotation, this extraction added a 100ms context window to both the start and end of each stop token. A subset of 55,628 stops (30,104 voiced, 25,524 voiceless) were manually annotated with stop burst presence or absence by the first author using a custom Praat script (v6.3 \cite{praat2023}). Approximately 5\% of the each corpus's voiced and voiceless stops were annotated, up to a maximum of 1,000 stops. Stops were annotated based on whether a separate stop closure and stop burst release can be observed \cite{csj-phone-doc}. To prepare for training, the stops were re-extracted with a 10ms context window, and 11k stops were pseudo-randomly selected (containing some stops from each corpus) as a validation set. To determine the relative amount of data needed to successfully train the classifier (RQ2), 6 subsets of the training data were created: \{500, 1000, 2000, 10k, 20k, 40k\}. For each of these data subsets, the same training procedure as Experiment 1 was applied, where each subset was trained using each of the 4 wav2vec2.0 base models.

\subsection{Results}

\begin{figure*}[t]
  \centering
  \includegraphics[width=\linewidth]{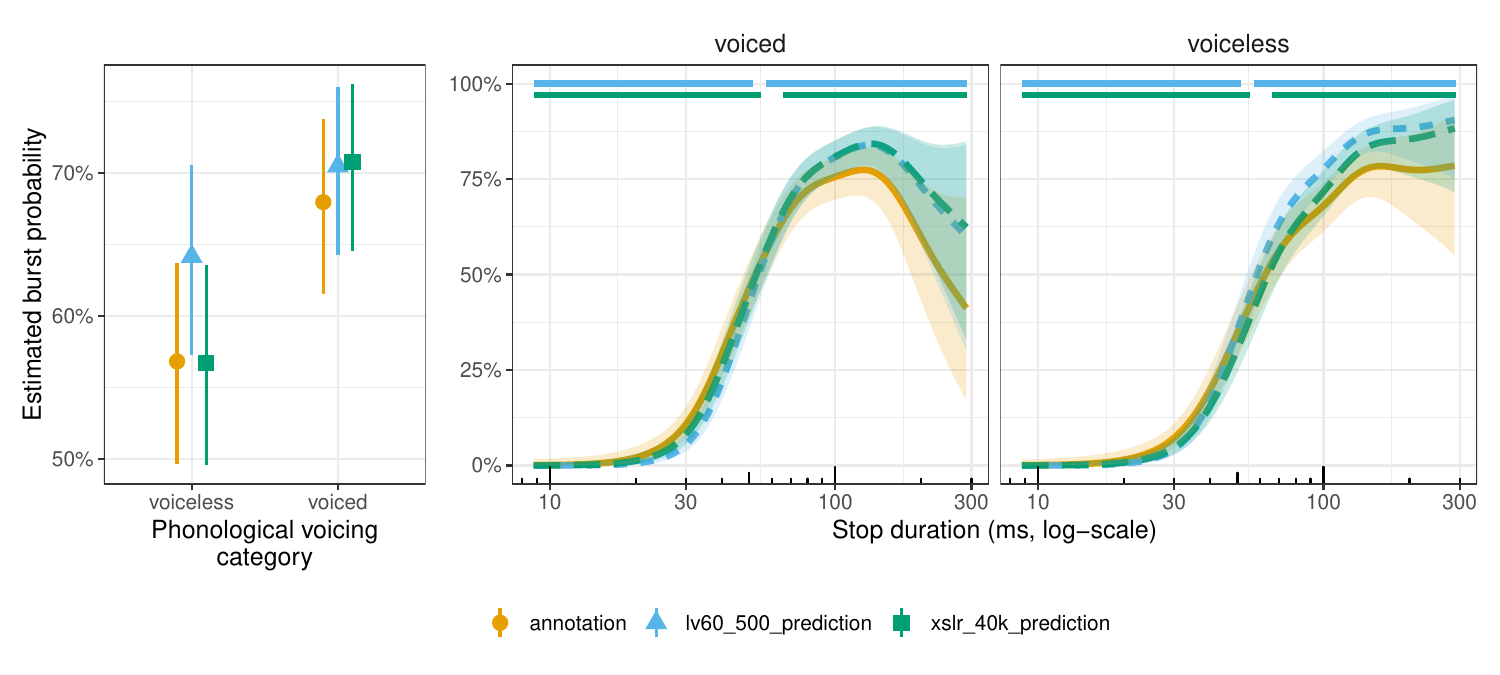}
  \caption{Model-estimated probability of a stop burst by phonological voicing category (left) and stop duration (right), separated by manual annotations (orange solid), predictions from the CSJ-C-pretrained \texttt{lv60} model trained on 500 SPADE tokens (blue short-dash), and predictions from the CSJ-C-pretrained \texttt{xslr} model trained with 40k SPADE tokens (green long-dash). Points and lines (left) and lines and shaded areas (right) indicate the mean estimate with 95\% confidence intervals. Bars on left indicate sections where the respective predictions significantly differ ($p < 0.05$) from hand annotations.}
  \label{fig:pred-plots}
\end{figure*}

Figure \ref{fig:ci_plot} shows the predictive accuracy and F1 for all models, evaluated on the SPADE validation set. For models trained on subsets of SPADE data (cols 1-6), we find that the amount of training data has a substantial effect on the predictive performance of the model. For all models trained with $\leq$1000 tokens (cols 1-2), accuracy is at chance level (with near-floor F1), indicating that these models failed to learn the acoustic dimensions of stop realisation. With between 2k and 10k training tokens (rows 3-4), performance varies between specific base models, with the models trained on the \texttt{lv60} and \texttt{xlsr} performing best. With at least 20k training tokens (rows 5-6), predictive accuracy appears to peak at 87-88\%. We also observe that a model not trained on the SPADE data (i.e. the CSJ-40k model) performs less well than a model trained on the same number of SPADE data samples (col 7), suggesting that models trained on manually-prepared clean speech data do not directly generalise to more variable, less-curated speech of a different language.

While high predictive performance can be achieved on highly variable speech, the large amount of training data required to achieve this performance ($\geqq$20k annotated stops) would place a practical barrier for use in most scenarios. To see whether high predictive performance could be achieved with a smaller training set, we explored the possibility of finetuning the SPADE data on top of the CSJ-C-trained models from Experiment 1, with the thinking that it might be possible to leverage the previously-learned acoustic properties of different stop realisations to `bootstrap' a better performance with a small amount of training data from the target distribution. The model training procedure for Experiment 2 was repeated (using the respective CSJ-C-trained model checkpoints as the initial training step), with the accuracy and F1 values reported in Figure \ref{fig:ci_plot} (cols 8-13). Compared with the performance from the SPADE-only trained models, models pre-trained with CSJ-C achieve much greater predictive performance with small training sets of 500-1k tokens (cols 8-9). While this is also only reached by models trained on the largest training sets (cols 12-13), models trained on all data subsets achieve accuracy and F1 scores similar ($\sim$1-2\%) of the 20k/40k models.

\section{Analysis of predicted stop patterns}
\label{sec:analysis}

Having demonstrated that wav2vec2.0 models can achieve good performance with both curated and non-curated data (RQ1, RQ2), we now consider the extent to which patterns from predicted stops follow those from hand-annotations (RQ3), where predictions from these models are used in a hypothetical study of variable stop realisation, using the set of $\sim$11k stops from the SPADE validation set. We consider the patterns of stop burst probability as a function of 1. phonological voicing and 2. stop duration, given the previous literature demonstrating these factors condition stop realisation \cite{Bouavichith2013,lavoie2001,warnertucker2011,katz2016,priva2020}.

Illustrating the difference in using more or less model training data, we chose to use the predictions from two CSJ-C $\rightarrow$ SPADE models representing `more' (\texttt{xslr-40k}) `less' (\texttt{lv60-500}) data, along with the manual annotations. Stop burst probability was modelled with a logistic generalised additive mixed model fit with the \texttt{mgcv} package (v1.8.42, \cite{wood2011}) in R v4.3 \cite{r2023}, with parametric predictors for the annotation type (\texttt{hand-annotation}, \texttt{lv60-500}, \texttt{xslr-40k}), and phonological voicing (\texttt{voiceless}, \texttt{voiced}), non-parametric (smooth) terms for log-transformed phone duration, and phone duration as a function of annotation type and phonological voicing, by-corpus random smooths, and by-speaker random slopes. Comparisons between annotation types were performed using the \texttt{emmeans} package (v1.9, \cite{emmeans2023}) and are reported as the estimated marginal difference between conditions ($\hat{\Delta}$), $t$-ratio, and $p$-value with Tukey adjustment for multiple comparisons.

Figure \ref{fig:pred-plots} illustrates the model-predicted stop burst probability for each of the annotation types \{hand annotation, \texttt{lv60-500}, \texttt{xslr-40k}\}. While we observe the reverse of the expected pattern for voicing (greater burst probability for voiced stops, Fig.~\ref{fig:pred-plots} left), we find that this pattern is shared across both the predicted and manual annotations. Comparing predictions for voiced and voiceless stops separately, we find that, for voiceless stops, the 500-stops model predicts the presence of a stop burst 6\% more often than both the hand annotations ($\hat{\Delta} = -0.31$, $t = -4.89$, $p < 0.0001$) and the 40k-model ($\hat{\Delta} = 0.31$, $t = 4.92$, $p < 0.0001$); the 40k-model and the annotations themselves do not differ in estimated burst probability ($\hat{\Delta} = 0.01$, $t = 0.08$, $p = 0.99$). For voiced stops, while the 40k-model is found to be significantly different from the hand annotations ($\hat{\Delta} = -0.13$, $t = -2.76$, $p = 0.01$), this corresponds to a $\sim$2\% difference in burst probability; this difference from the hand annotations was also found for the 500-model, though was not significantly different ($\hat{\Delta} = -0.12$, $t = -2.24$, $p = 0.07$). We also find that the effect of stop duration on burst likelihood also differs by annotation type (Figure \ref{fig:pred-plots}, right), where a nested model not containing the by-annotation smooth terms results in a worse fit to the data (${\chi}^2(4) = 114.21$, $p < 0.0001$). For shorter stops, this corresponds to small differences in probability (e.g. $\sim$1\% at 20ms), while the probability is overestimated by approximately 7-8\% for longer stops (e.g. $\geq$100ms), with both prediction types following similar patterns.

\section{Discussion}
\label{sec:discussion}

The goal of this study was to determine whether self-supervised speech models, such as wav2vec2.0, can be utilised as a tool for automatically annotating the variable realisation of stops, to allow for both the study of variable lenition in its own right \cite{lavoie2001,katz2016}, and further downstream annotations. Specifically, we are interested in whether such models can predict the presence or absence of a stop burst in datasets reflecting two broad types of collected speech data: 1. cleanly recorded speech data with manual correction of segmental labels and boundaries, and 2. more opportunistically-collected speech data that vary in their recording quality, which have not undergone any manual correction following forced alignment. After demonstrating that wav2vec2.0 models can achieve high performance on highly-curated Japanese spontaneous speech (Sec.~\ref{sec:exp1}), we explore the same training procedure for a large multi-corpus collection of force-aligned English speech (Sec.~\ref{sec:exp2}). Burst detection accuracy reaches 88\% for this data, though a large amount of manual annotations ($\geq$20k) is required to achieve this. By using models pre-trained with CSJ-C data, however, similar accuracy can be achieved with 500-1k tokens. Comparing the patterns of predicted burst realisation with hand-annotations (Sec.~\ref{sec:analysis}), we find that predicted burst presence largely follows the same patterns as those from annotations for both phonological voicing and stop duration; while statistically significant differences are observed between predictions and the annotations, the magnitude of these differences are relatively small compared to broader patterns (e.g. across the range of stop durations). 

These results demonstrate that wav2vec2.0 can be trained to accurately detect the presence of stop bursts in spontaneous speech data, across multiple languages and in cases where the target data is highly variable in its recording quality and data curation. By using models pre-trained for classifying stop realisation, high accuracy could be achieved with a very small number of annotated tokens. This lowers the barrier for the annotation of a speech corpus, both in requiring minimal labour for annotation, as well as for cases where the corpus is not large enough to support the annotation of thousands of tokens.\footnote{Software \& models are available at \url{https://github.com/james-tanner/wav2vec-burst-detection}.}

\section{Acknowledgements}
The authors thank the SPADE Data Guardians, Rachel Macdonald, Michael McAuliffe, and Vanna Willerton. Computational resources were provided by the Digital Research Alliance of Canada. This research was supported by a T-AP Digging into Data award in the form of the following grants: ESRC Grant \#ES/R003963/1, NSERC/CRSNG Grants \#RGPDD-501771-16 and \#RGPIN-2023-04873, SSHRC/CRSH Grant \#869-2016-0006, and NSF Grant \#SMA-1730479. This research was also supported by a Canada Research Chair \#CRC-2023-00009 (MS) and a British Academy Postdoctoral Fellowship (JT).

\bibliographystyle{IEEEtran}
\bibliography{interspeech_stops_paper}

\end{document}